\documentclass[runningheads]{llncs}
\usepackage[T1]{fontenc}
\usepackage{orcidlink}
\usepackage{marvosym}
\usepackage{graphicx}
\begin{document}
\title{Deformable Image Registration with Multi-scale Feature Fusion from Shared Encoder, Auxiliary and Pyramid Decoders}
%
%
\author{Hongchao Zhou\orcidlink{0009-0006-1883-4435} \and
	Shunbo Hu \textsuperscript{(\Letter)}\orcidlink{0000-0002-1442-0976} }

\authorrunning{Zhou et al.}
%
\institute{School of Information Science and Engineering, Linyi University, Lin Yi, 276000, Shandong, China \\
	\email{hushunbo@lyu.edu.cn}
}
\maketitle              
\begin{abstract}
	In this work, we propose a novel deformable convolutional pyramid network for unsupervised image registration. Specifically, the proposed network enhances the traditional pyramid network by adding an additional shared auxiliary decoder for image pairs. This decoder provides multi-scale high-level feature information from unblended image pairs for the registration task. During the registration process, we also design a multi-scale feature fusion block to extract the most beneficial features for the registration task from both global and local contexts. Validation results indicate that this method can capture complex deformations while achieving higher registration accuracy and maintaining smooth and plausible deformations.
	
\end{abstract}
\section{Introduction}
Deformable image registration (DIR) is a basic task in the field of computer vision, which is widely used in medical diagnosis, surgical guidance, disease detection and so on, with the goal of determining an optimal spatial transformation that warps the moving image and align it with the corresponding fixed image. Traditional methods \cite{klein2009elastix,klein2009evaluation} usually treat DIR as an optimization task, trying to minimize the energy function in an iterative manner, but it requires a lot of computation and long processing time.

These limitations have spurred the rapid development of deep learning-based DIR methods. Recently, deep learning DIR methods, especially unsupervised approaches \cite{chen2024transmatch,zheng2024residual}, have gained significant attention due to their ability to operate without any labeled data. Pyramid network \cite{ma2024iirp,wang2024recursive} have shown exceptional performance in recent unsupervised methods. Specifically, in a pyramid network, the moving and fixed images are typically encoded separately with shared parameters. Their multi-scale features are then sent to the decoder to achieve coarse to fine registration.

In this work, we propose a novel pyramid network-based DIR method. Firstly, we recognize that independent feature extraction can provide distinctive information for registration, enabling the network to identify correspondences between image pairs. Building on this, the decoder of our method is similar to these in papers \cite{ma2024iirp,wang2024recursive}, and we add a shared auxiliary decoder for each of the moving and fixed images. This decoder's primary function is to provide multi-scale high-level features of different images for registration. These additional high-level features enhance the network's understanding of image details and global structures, contributing to more precise registration. Secondly, we design a multi-scale feature fusion block (MSFB) that receives inputs from the encoder, the auxiliary decoder, and the coarse deformation field obtained from the previous registration scales. The MSFB selects the most beneficial features for registration from both global and local perspectives by global and local attention mechanism, effectively filtering out redundant information and retaining key features. Finally, we validate our method on the dataset from LUMIR task in Learn2Reg 2024. Our method achieves better registration accuracy and smoother, more plausible deformations compared to several learning-based DIR methods. 

\section{Method}
The proposed method mainly consists of a shared encoder, a shared auxiliary decoder and a fusion pyramid decoder. The fusion pyramid decoder includes five scales. Except for the smallest scale, each scale is constructed by MSFB. The pyramid decoder first merges the smallest scale low-level features and predicts the coarse DDF. This DDF is then upsampled and applied to the moving feature map at the current scale. The moving feature map, the fixed feature map, and the high-level moving and fixed features from the shared auxiliary decoder at the same scale are input into the MSFB. This process removes redundant information and produces fused features, predicting the DDF at the current scale. The subsequent process is the same as the current scale, with the final refined DDF obtained by iteratively combining the coarse DDFs from each scale.

In our method, we choose to use normalized cross-correlation $\mathcal{L}_{ncc}$ \cite{rao2014application} and deformation regularization $\mathcal{L}_{reg}$ \cite{balakrishnan2019voxelmorph} to train the network. Thus, the total loss $\mathcal{L}$ is expressed as:
\begin{equation}
	\mathcal{L}=\mathcal-(\alpha{L}_{ncc}(I_{f},I_{m}\circ \phi) + \beta{L}_{ncc}(I_{f},I_{m}\circ \hat{\phi} ) )+\mathcal\lambda {L}_{reg}(\phi ), \label{eq1}   
\end{equation}
where $\hat{\phi}$ is obtained by sampling the DDF at the scale of 80$\times$112$\times$96. We first calculate the deformation using $\hat{\phi}$. This approach optimizes the auxiliary decoder during back propagation and provides the fusion pyramid decoder with deformation-included feature information, leading to more accurate deformations. The parameters $\alpha$, $\beta$ and $\lambda$ are hyperparameters. In the experiments, we set $\alpha$ to 0.7, $\beta$ to 0.3, and $\lambda$ to 1.

\section{Results}
\begin{figure}[!t]
	\centerline{\includegraphics[width=1.0\textwidth]{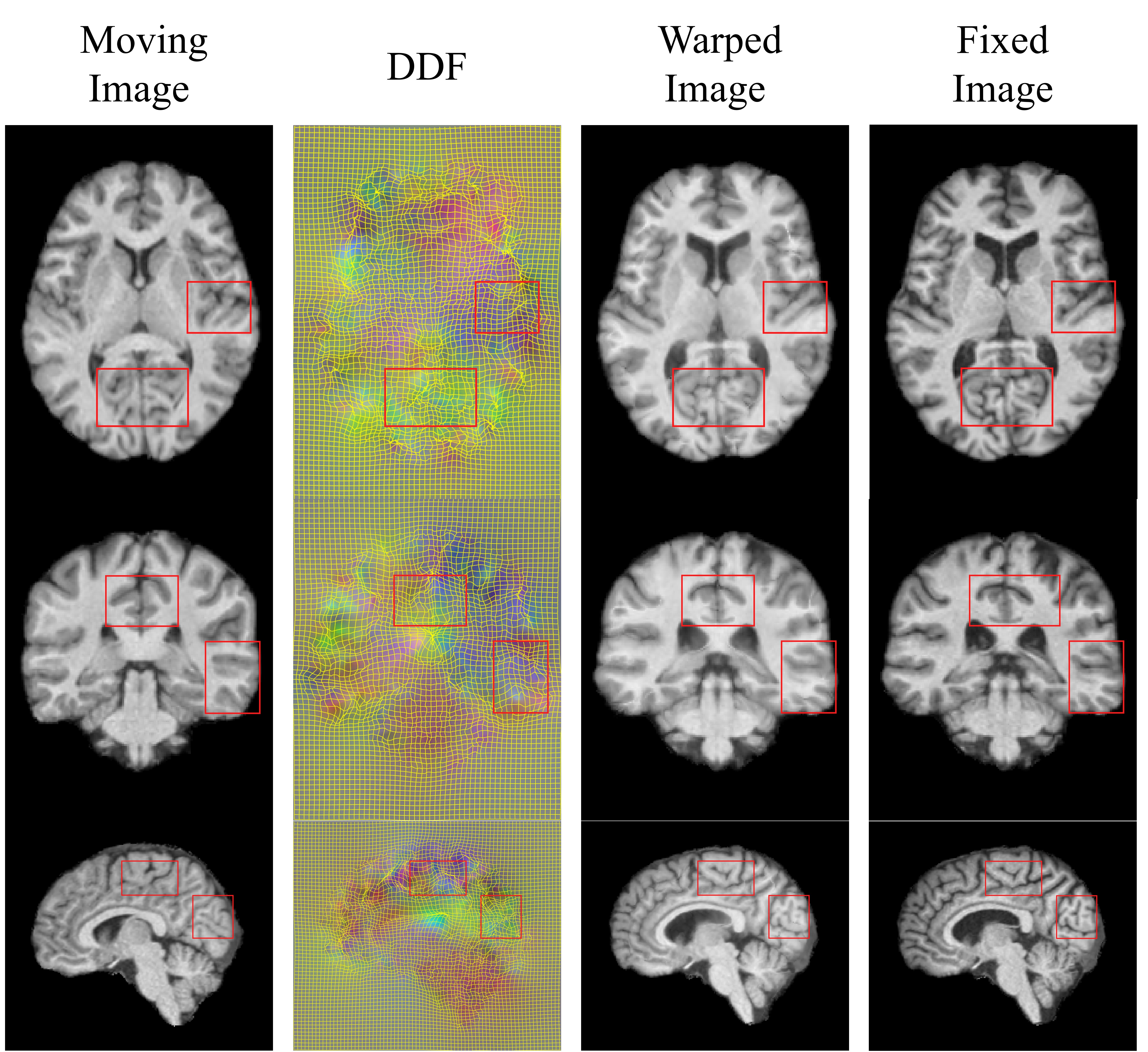}}
	\caption{An example of our method results in registration. From left to right are moving image, deformation field, warped image and fixed image. The red box highlights the significant areas.
	} \label{fig1}
\end{figure}
\subsubsection{Dateset and preprocessing}
We evaluate our method by performing registration on the dataset from the LUMIR task \cite{dufumier2022openbhb,marcus2007open,taha2023magnetic} in the Learn2Reg challenge. The dataset contains 3384 training images. For each calculation, we randomly select two images as the moving and fixed images, resulting in a total of 1692 image pairs. Additionally, a validation set consisting of 40 images is provided, forming 38 pairs for validation purposes. All images are converted to NIfTI, resampled, and cropped to size of 160$\times$224$\times$192 with a voxel spacing of 1$\times$1$\times$1 mm$^{3}$. 
\begin{table}
	\centering
	\caption{Quantitative evaluation results on the Learn2Reg challenge dataset. Dice, TRE and HdDist95 represents registration accuracy. NDV is used to describe whether the deformation is smooth and plausible.}\label{tab1}
	\begin{tabular}{|c|c|c|c|c|}
		\hline
		Method     & Dice $\uparrow$                                    & TRE $\downarrow$ (mm) & NDV $\downarrow$ (\%) & HdDist95 $\downarrow$ \\ \hline
		Affine     & 0.5657 ± 0.0263                        & 4.3543  & 0      & 4.7876   \\
		deedsBCV   & 0.6977 ± 0.0274                        & 2.223   & 0.0001  & 3.954    \\
		SyN        & 0.6988 ± 0.0561                        & 2.6497  & 0       & 3.7048   \\
		VoxelMorph & 0.7186 ± 0.0340                        & 3.1545  & 1.1836  & 3.9821   \\
		SynthMorph & 0.7243 ± 0.0294                        & 2.6099  & 0       & 3.573    \\
		TransMorph & 0.7594 ± 0.0319                        & 2.4225  & 0.3509  & 3.5074   \\
		Ours        &  0.7727 ± 0.0276 & 2.4663  & 0.0130  & 3.3197   \\ \hline
	\end{tabular}
\end{table}
\subsubsection{Evaluation criteria}
To assess registration performance, we utilize the Dice score (Dice), Target registration error (TRE) and 95\% Hausdorff Distance(HdDist95). Higher Dice , lower TRE and lower HdDist95 indicate better registration precision. Additionally, folding is assessed by the Non-diffeomorphic volumes (NVD) \cite{liu2024finite} on DDFs, with lower NVD values reflecting smoother and more plausible deformations.

\subsubsection{Experiments}
We compare our method with five baseline DIR methods, encompassing two tradition methods(SyN and deedsBCV) and three deep learning methods (VoxelMorph, SynthMorph and TransMorph). Table~\ref{tab1} presents the quantitative evaluation results for all methods on the Learn2Reg challenge dataset. Our method achieve the highest registration accuracy and produce smoother and more plausible deformations compared to the two mainstream unsupervised methods, VoxelMorph and TransMorph. Fig.~\ref{fig1} shows examples of registration results from our method on the Learn2Reg challenge validation set. The areas within the red boxes indicate regions of significant deformation during registration. Our method demonstrates excellent performance in this region.

\section{Discussion}
Our method extends the traditional pyramid network by leveraging multi-scale fusion feature information from three sources: low-level encoder, high-level auxiliary decoder, global and local fusion pyramid decoder, which enhances the network's understanding of image details and global structures. Additionally, our designed MSFB effectively filters out redundant information and retains key features from both global and local perspectives, further improving registration performance. 

Experiments show that our method effectively handles large deformation issues. Additionally, due to our MSFB, the fusion features provide more accurate detail information for the registration task, allowing our method to refine the registration across different scales and achieve higher performance. Overall, our method not only enhances registration performance but also ensures smooth and plausible deformations.
%
%
%
%

\bibliography{ref.bib}
\bibliographystyle{splncs04}
\end{document}